\pdfoutput=1

\documentclass[11pt]{article}

\usepackage[preprint]{acl}

\usepackage{times}
\usepackage{latexsym}

\usepackage[T1]{fontenc}

\usepackage[utf8]{inputenc}

\usepackage{microtype}

\usepackage{inconsolata}

\usepackage{graphicx}
\usepackage{subcaption}
\usepackage{multirow}
\usepackage{multicol}
\usepackage{booktabs}

\title{From Dense to Dynamic: Token-Difficulty Driven MoEfication of Pre-Trained LLMs}

\usepackage[textwidth=2.5cm]{todonotes}

\author{
\textbf{Kumari Nishu}, \textbf{Sachin Mehta}\thanks{Contributed when employed by Apple.}, \textbf{Samira Abnar}, \textbf{Mehrdad Farajtabar} \\ \textbf{Maxwell Horton}, \textbf{Mahyar Najibi}, \textbf{Moin Nabi}, \textbf{Minsik Cho}, \textbf{Devang Naik}
\\
Apple
\\
}

\usepackage{amssymb, amsmath, mathtools, bm}
\def\R{\mathbb{R}}
\def\bx{\bm{X}}
\def\by{\bm{Y}}
\def\L{\mathcal{L}}

\newcommand{\win}{\bm{W}^{(IN)}}
\newcommand{\wout}{\bm{W}^{(OUT)}}
\newcommand{\bp}[1]{\left( #1 \right)}

\newcommand{\ip}[2]{\left< #1, #2 \right>}

\DeclarePairedDelimiter\floor{\lfloor}{\rfloor}
\DeclareMathOperator{\mname}{DynaMoE}

\begin{document}
\maketitle
\begin{abstract}

Training large language models (LLMs) for different inference constraints is computationally expensive, limiting control over efficiency-accuracy trade-offs. Moreover, once trained, these models typically process tokens uniformly, regardless of their complexity, leading to static and inflexible behavior. In this paper, we introduce a post-training optimization framework, $\mname$, that adapts a pre-trained dense LLM to a token-difficulty-driven Mixture-of-Experts model with minimal fine-tuning cost. This adaptation makes the model dynamic, with sensitivity control to customize the balance between efficiency and accuracy. $\mname$ features a token-difficulty-aware router that predicts the difficulty of tokens and directs them to the appropriate sub-networks or experts, enabling larger experts to handle more complex tokens and smaller experts to process simpler ones. Our experiments demonstrate that $\mname$ can generate a range of adaptive model variants of the existing trained LLM with a single fine-tuning step, utilizing only $10B$ tokens, a minimal cost compared to the base model's training. Each variant offers distinct trade-offs between accuracy and performance. Compared to the baseline post-training optimization framework, Flextron, our method  achieves similar aggregated accuracy across downstream tasks, despite using only $\frac{1}{9}\text{th}$ of their fine-tuning cost.





\end{abstract}

\section{Introduction}


 Large language models (LLMs) have significantly advanced the field of natural language processing, showcasing strong capabilities in addressing complex tasks \cite{LLM1, LLM2_LLama, LLM3_Wei2022ChainOT}. However, their large size presents challenges, particularly in terms of high memory and computational demands, which can limit their deployment in resource-constrained settings. To address this, LLMs must be optimized for specific memory and computational constraints \cite{Touvron2023Llama2O}. However, designing multi-billion-parameter models for every use case is not cost-effective, as it demands substantial training time, data, and resources.

Some prior works have focused on adapting large LLMs for resource-constrained use cases by distilling knowledge from larger models into smaller ones~\citep{hsieh-etal-2023-distilling} or pruning model parameters to reduce computational demands~\citep{sun2024a}. While these methods effectively enable the use of large LLMs in low-resource scenarios, they often lead to performance degradation and require careful balancing between efficiency and accuracy. Alternatively, other approaches have investigated many-in-one LLM designs, MatFormer~\citep{Devvrit2023MatFormerNT} and SortedNet~\citep{Sortednet}, to employ multiple sub-networks within a single model to accommodate different computational budgets. These architectures use nested structures integrated into the standard LLM framework. However, they require non-standard methodologies and significantly longer, more resource-intensive training processes, which can offset the intended efficiency benefits.

Mixture-of-Experts (MoE) models~\citep{shazeer2017,Du2021GLaMES,fedus_switch,zoph2022st,he2024mixture} have emerged as a promising alternative to dense models, offering improved efficiency by sparsely activating select sub-modules or experts. This selective activation enables MoEs to achieve high performance while using fewer computational resources during inference. However, training MoEs from scratch remains resource-intensive and each expert becomes static, often requiring fixed compute budget irrespective of the input complexity.

\begin{figure*}[!t]
    \centering
    \includegraphics[width=0.95\textwidth]{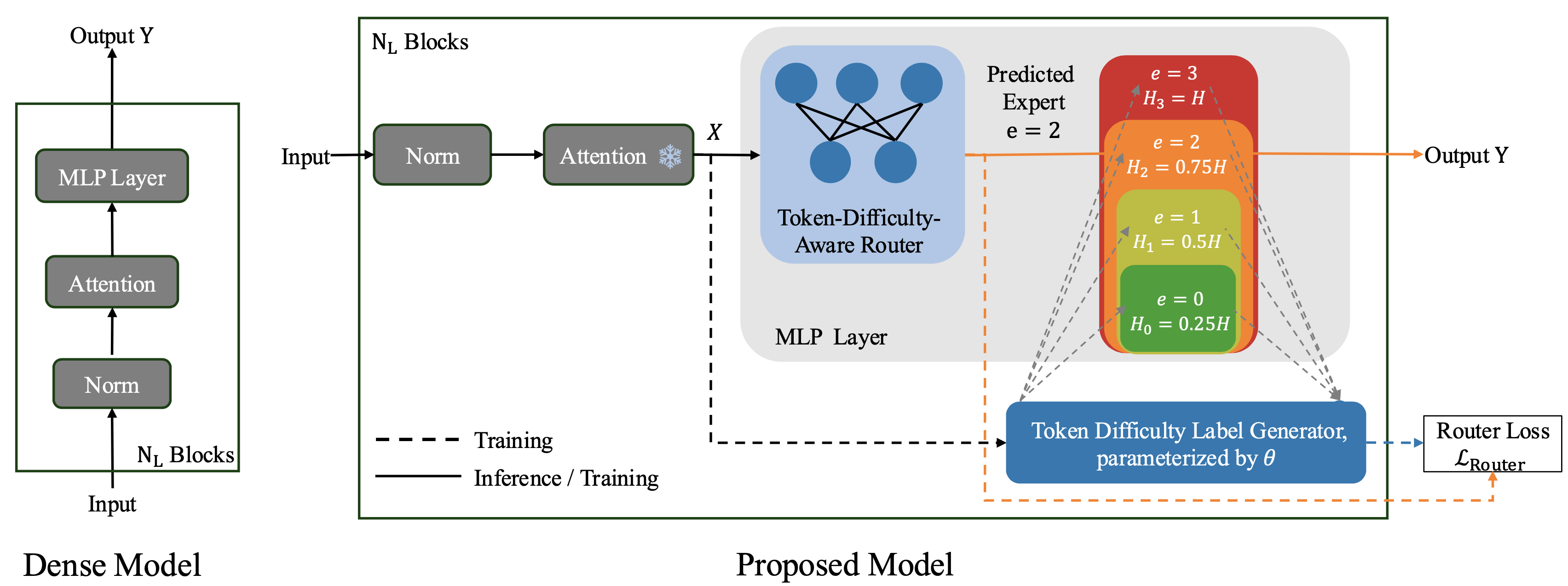} 
    \caption{Overview of our proposed post-training optimization framework, $\mname$. The left part represents the base pre-trained LLM, while the right part shows the adapted $\mname$ model.
    }
    \label{fig:overview}
\end{figure*}

Flextron~\citep{Cai2024FlextronMF} explored a post-training methodology by integrating the MoE concept into a nested elastic structure within the MLP layers, creating heterogeneous experts of different sizes, selected by a router conditioned on the input data. However, the lack of supervision in the router training leads to sub-optimal input complexity adaptation. Furthermore, the router lacks a parameter to customize its sensitivity to token complexity, limiting its flexibility and performance in handling diverse use-cases. \citet{Salehi2023SHARCSET} proposed an input-adaptive approach that predicts the difficulty of input data and dynamically adjusts the network's width accordingly. In the absence of ground-truth difficulty labels, they relied on heuristic methods for label generation, which may limit precision and consistency in difficulty estimation.

To address their shortcomings, we introduce $\mname$, a post-training optimization framework designed to transform a dense LLM into a token-difficulty-driven MoE model. $\mname$ leverages the insight that not all tokens require the full capacity of a model's weights. For example, in the sentence ``Geoffrey did his PhD at the university of Edinburgh'', simpler tokens like ``at the university of'' are predictable using prior context, while more complex tokens like "Edinburgh" demand broader contextual understanding. To maximize efficiency, $\mname$ selectively activates nested sub-components of the MLP, referred as experts, based on the predicted difficulty of each token. To this end, we make the following contributions:

\begin{itemize}
    \item The framework includes a novel token-difficulty-aware router, trained to predict token hardness and assign it to the appropriate expert dynamically.
    
    \item Due to the lack of ground truth notion of hardness, we introduce a method to derive token difficulty labels which serve as supervision signals for training the router. This approach allows a token to have varying difficulty labels across different layers.

    \item A simplified post-training optimization framework that efficiently adapts a pre-trained dense LLM into a token-difficulty-driven MoE model, featuring a sensitivity parameter to customize the efficiency vs accuracy trade-off.

\end{itemize}

\section{Method}
In this section, we describe our proposed post-training optimization framework, $\mname$, which transforms a dense LLM into an MoE model for adaptive inference based on token difficulty. The process involves three key steps: $(1)$ defining heterogeneous experts by splitting the MLP layers of the dense LLM; $(2)$ generating token labels during training to represent token difficulty; and $(3)$ training a router to predict token difficulty while fine-tuning the model. We detail these steps in the below sub-sections.

\subsection{Defining Heterogeneous Experts}
In this work, we focus on defining experts into the MLP layers of the LLM \cite{Devvrit2023MatFormerNT}, as these layers account for the majority of the compute and operate on a token-by-token basis. The overview of $\mname$ is depicted in Fig. \ref{fig:overview}. The left part of the figure denotes the base pre-trained model which consists of the normalization layers, attention layers and the MLP layers in each transformer block. The right part shows the adapted $\mname$ model, where the original single MLP layer is transformed into multiple nested FFN blocks or experts. Such expert formation introduces no additional parameters to the base model, aside from the router. This design draws inspiration from adaptive width reduction in transformer \cite{Salehi2023SHARCSET} and recent works like Matformer \cite{Devvrit2023MatFormerNT} and Flextron \cite{Cai2024FlextronMF}. The attention layers remain frozen, and the MLP layers adapted to nested experts are fine-tuned in $\mname$.

Let $D$ and $H$ denote the embedding and the hidden dimensions of the MLP layer respectively. The input to the MLP layer is $\bx \in \R^{B \times D}$ and the output is $\by \in \R^{B \times D}$, where $B$ is the batch dimension. The MLP layer with two fully connected layers is represented by weight matrices $\win \in \R^{H \times D}$ and $\wout \in \R^{D \times H}$. In order to get best results, we first rearrange these fully-connected layers,  $\win$ and $\wout$, to have the most important rows/columns in the beginning of the matrix so that they can be included in all of the experts \cite{Samragh2023WeightSD}. There are a total of $E$ experts indexed using $e \in  \{0, 1, \ldots, E-1\}$. Each expert gets a portion $H_e$ of the weight matrices $\win$ and $\wout$, sliced over the hidden dimension $H$. The value $H_e$ is obtained as a fraction of $H$ as, 
\begin{equation}
    H_e = \floor*{\bp{\frac{e+1}{E}}\cdot H},
\end{equation}
consequently, $H_0 < H_1 < \cdots < H_{E-1}$ and $H_{E-1}=H$. Note that the expert with index $E-1$ utilizes the full MLP layer. The restriction of the matrices $\win$ and $\wout$ to the expert width $H_e$ is obtained using the slicing operator that selects the first $H_e$ rows and columns respectively as 
\begin{align}
    \win_e &= \win[0:H_e,\, :], \\
    \wout_e &= \wout[:,\, 0:H_e].
\end{align}
    
With $\sigma$ as the activation function, the output $\by_e$ of the MLP layer corresponding to the expert $e$ can thus be obtained as, 
\begin{equation}\label{eq:expert}
    \by_e = \sigma \bp{ \bx \cdot \bp{ \win_e}^T  } \cdot \bp{ \wout_e}^T.
\end{equation}

\begingroup
    \renewcommand{\arraystretch}{1.5}
    \begin{table*}[!t]
        \centering
        \resizebox{\textwidth}{!}{%
        \begin{tabular}{c|c|c|c|c|c|c|c|c|c|c|c}
            \toprule
            & \textbf{Cost (\#Tokens)} & \textbf{Params} & \textbf{ARC-e} & \textbf{LAMBADA} & \textbf{PIQA} & \textbf{WinoGrande} & \textbf{Avg4} & \textbf{SciQ} & \textbf{HellaSwag} & \textbf{ARC-c}  & \textbf{Avg7} \\
            \toprule
            Base Mistral 7B & - & 7B & 80.2 & 75.1 & 80.8 & 75.5 & \textbf{77.8} & 96.4 & 61.4 & 50.5  & 74.2 \\
             
            \hline
            $\mname$ $\theta=0.9$ & \textbf{10B}  & 6B & 75.0 & 71.0 & 78.3 & 71.8 & \textbf{74.0} & 95.2 & 57.9 & 41.5 & 70.1 \\
            
            $\mname$ $\theta=0.8$ & \textbf{10B}  & 5.1B & 69.9 & 68.0 & 78.0 & 66.0 & \textbf{70.5} & 94.2 & 54.5 & 35.2 & 66.5 \\
            
            
           $\mname$ $\theta=0.7$ & \textbf{10B}  & 4.6B & 66.0 & 65.9 & 75.4 & 63.4 & 67.7 & 93.6 & 52.1 & 31.7 & 64.0 \\
            
           \hline
        Base Llama2-7B $^\dagger$ & -  & 6.5B & 75.1	& 71.5	& 77.5 &	69.1 & 73.3 &  &  &  &  \\
        
        Flextron $^\dagger$ & 93.57B & 4.1B & 68.6	& 65.1	& 76.1 &	63.7 & 68.3 &  &  &  &  \\
            \bottomrule
        \end{tabular}
        }
        \vspace{0.1 in}
        \caption{Evaluation of $\mname$ models with different sensitivity factor $\theta$ on downstream tasks, using 0-shot non-normalized accuracy metric. Our base model is Mistral 7B \citep{Jiang2023Mistral7}. $(^\dagger)$: results from Flextron \citep{Cai2024FlextronMF} used as our baseline. \textit{Params} denotes the average number of total activated parameters, aggregated over the downstream tasks. \textit{Avg4} averages over \textit{ARC-e, LAMBDA, PIQA, WinoGrande}, while \textit{Avg7} averages over all tasks.
        }
        \label{tab:result}
    \end{table*}
\endgroup

\subsection{Generating Token Difficulty Label}\label{subsec:generate_gt_label}


We aim to train a token-difficulty-aware router to dynamically assign tokens to an appropriate expert. But there is no ground-truth label denoting token difficulty to train such a router. To this end, we propose a method to estimate the token difficulty and generate a derived-ground-truth difficulty label during training. This is shown  as ``Token Difficulty Label Generator'' in Fig. \ref{fig:overview}. 

First, we pass the input to all experts and generate the output $\by_e$ for each $e \in [E]$. Then, for each token $b \in [B]$ and each expert $e \in [E]$, we compute a similarity score $S_{b,e}$ that measures how similar is the output of the expert $e$ compared to the output of the full MLP layer $e={E-1}$ for that token. We calculate this similarity as,
\begin{equation}
    S_{b,e} = \frac{\ip{\by_e[b,\,:]}{\by_{E-1}[b,\,:]}}{\ip{\by_{E-1}[b,\,:]}{\by_{E-1}[b,\,:]}}.
\end{equation}
Here, $\ip{\cdot}{\cdot}$ denotes the dot-product between two vectors. We use dot-product in this calculation as it accounts for both the magnitude and the direction of the tensors being compared.

Finally, we generate a derived ground-truth hardness label $l_b$, representing the target expert index for token $b$. Given a threshold $\theta$, we assign $l_b$ as the smallest expert index $e$ satisfying $S_{b,e} > \theta$, that is, $l_b=\min\{e \in [E] \mid S_{b,e}>\theta\}$. We say that a token is easier if it has a smaller label $l_b$, that is the similarity score for a smaller expert is higher than threshold $\theta$. In such cases, processing the token with the smaller expert incurs less compute without much compromise in the accuracy. During the forward pass of fine-tuning of $\mname$, we generate the token difficulty labels. These labels are then used in the backward pass to compute the router loss, which trains the router.

\subsection{Training a Token-Difficulty-Aware Router}
The output of a router is in $\R^{B \times E}$, denoting logits over the $E$ experts. Each router is parameterized by two linear layers, projecting the token embedding from dimension $D$ to $U$ and subsequently to $E$. In our experiments, we use $U=256$, resulting in total parameters added by the routers across all layers to $33.6M$, which is only $0.51\%$ of the base model size. 

We train the router using the derived labels from Section~\ref{subsec:generate_gt_label}. The objective is to learn the expert prediction using the derived-ground-truth labels to mimic token assignment based on their complexity and need. Hence, we impose the cross-entropy loss on the router to guide to this behavior and call it as the router loss. The overall objective function of $\mname$ is given as,
\begin{equation}\label{eq:loss}
    \L = \lambda_{LLM} \cdot \L_{LLM} + \lambda_{Router} \cdot \L_{Router}.
\end{equation}
Here, $\L_{LLM}$ is the main LLM Cross-entropy loss and $\L_{Router}$ is the router loss.  $\lambda_{LLM}$ and $\lambda_{Router}$ are hyper-parameters, denoting the weights of the respective losses.

\section{Experimental Set up}
\subsection{Model and Dataset}
\textbf{Model}: $\mname$ provides a simplified post-training approach to convert any dense LLM to an MoE model with a tunable sensitivity factor $\theta$, the similarity threshold, to achieve desired latency reduction and the tolerance for drop in accuracy. $\mname$ integrates seamlessly with any transformer model, regardless of the architecture. To showcase the effectiveness of the method, we use Mistral 7B model \cite{Jiang2023Mistral7}, a widely-used open-source pre-trained language model, as the base model.

\textbf{Dataset}: For $\mname$ fine-tuning, we use a small subset (10B tokens) of the Falcon RefinedWeb dataset \cite{refinedweb}. Falcon RefinedWeb is an open-source dataset which contains high-quality web data. This minimal fine-tuning overhead enables a cost-effective conversion of any pre-trained LLM into an MoE variant for faster inference.

\subsection{Training Details}
We first reorder the pre-trained weight matrices \cite{Samragh2023WeightSD} in the MLP blocks before fine-tuning $\mname$ so that the most important weights can be included in all the experts. We run the dense LLM on a subset of the data and collect the absolute activations from the MLP layer, $Y \in R^{B,H}$. This subset can be a very small portion of the training data, $0.004\%$ tokens of the Falcon RefinedWeb in our case. Subsequently, we aggregate the activations along the batch dimension and across all the subset samples to obtain an importance score for each neuron in the hidden dimension. Using this importance score, we sort the MLP matrices.

 Next, we fine-tune the $\mname$ model using only $10B$ tokens with AdamW optimizer \cite{Loshchilov2017DecoupledWD} and a fixed learning rate of $10^{-5}$. We keep the attention layers frozen. We set $\lambda_{LLM}$ to $0.2$ and $\lambda_{Router}$ to $1$ in Equation~\eqref{eq:loss}, the objective function for fine-tuning. We experiment with different values of threshold $\theta \in {\{0.7, 0.8, 0.9\}}$ to build a family of $\mname$ models with varying sensitivity parameter. A low sensitivity parameter, that is a smaller value of $\theta$, makes the system less reactive, favoring smaller experts for most tokens and only escalating to bigger experts for significantly complex tokens. And a high sensitivity parameter makes the system more reactive, escalating to bigger experts even for moderately complex tokens. We use $4$ experts ($E=4$) with sizes $0.25H$, $0.5H$, $0.75H$, and $H$ respectively. We denote the size of expert with index $e$ as $H_e$.

\begin{figure*}[!t]
      \centering
     \begin{subfigure}[b]{0.32\linewidth}
         \centering
         \includegraphics[width=\linewidth]{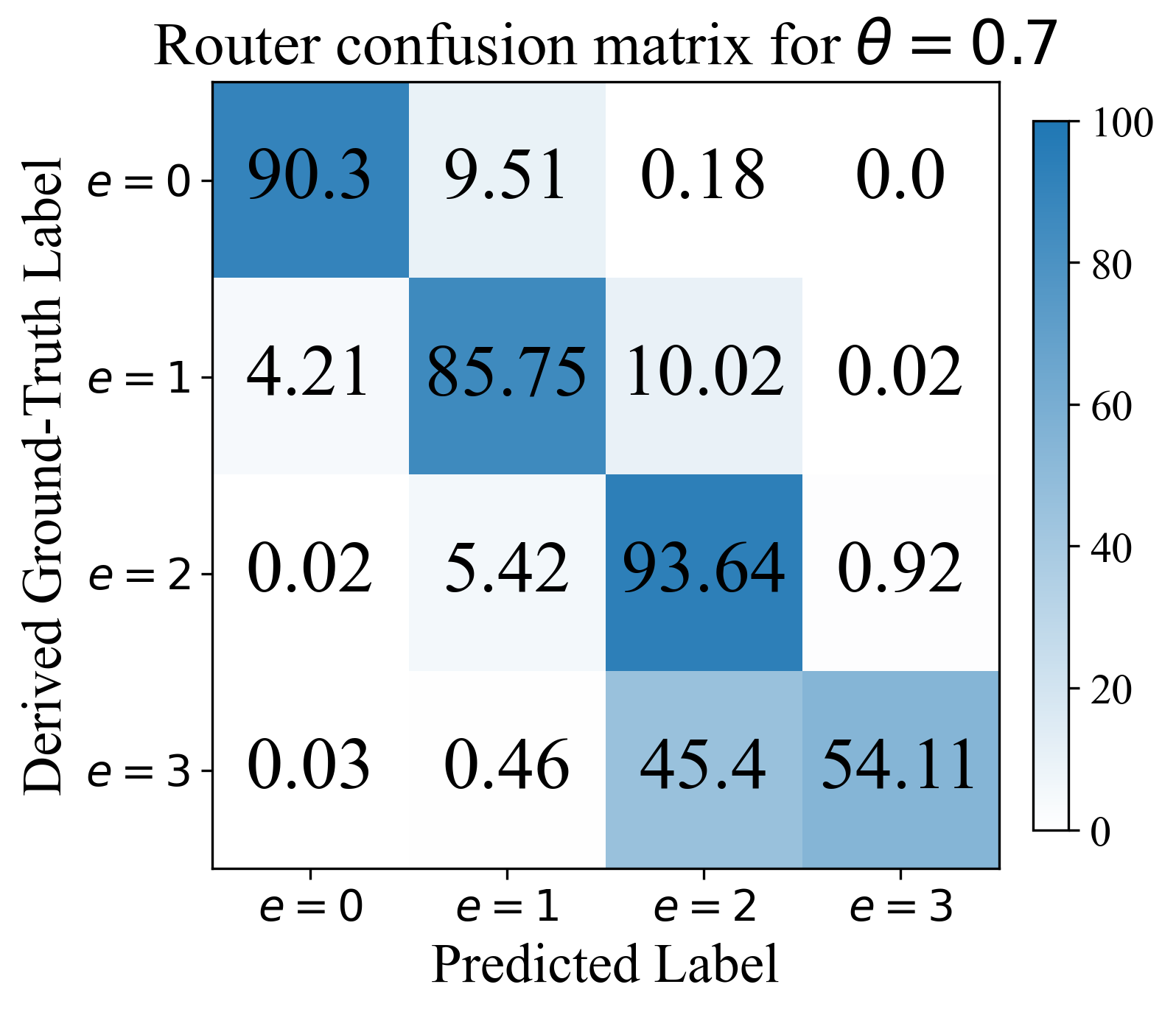}
         \caption{}
         \label{fig:0.7_routing}
     \end{subfigure}
    \begin{subfigure}[b]{0.32\linewidth}
         \centering
         \includegraphics[width=\linewidth]{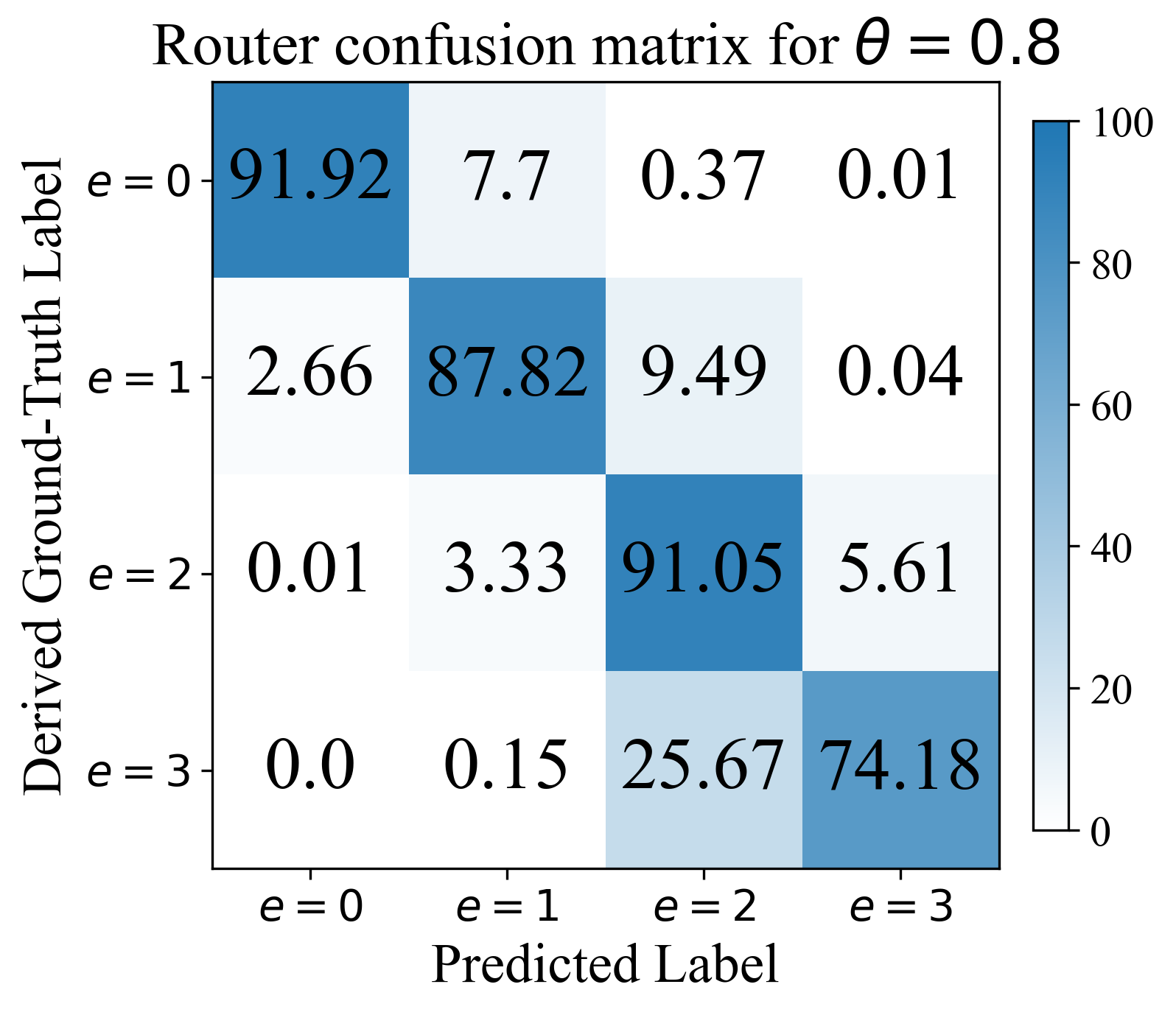}
         \caption{}
         \label{fig:0.8_routing}
     \end{subfigure}
    \begin{subfigure}[b]{0.32\linewidth}
         \centering
         \includegraphics[width=\linewidth]{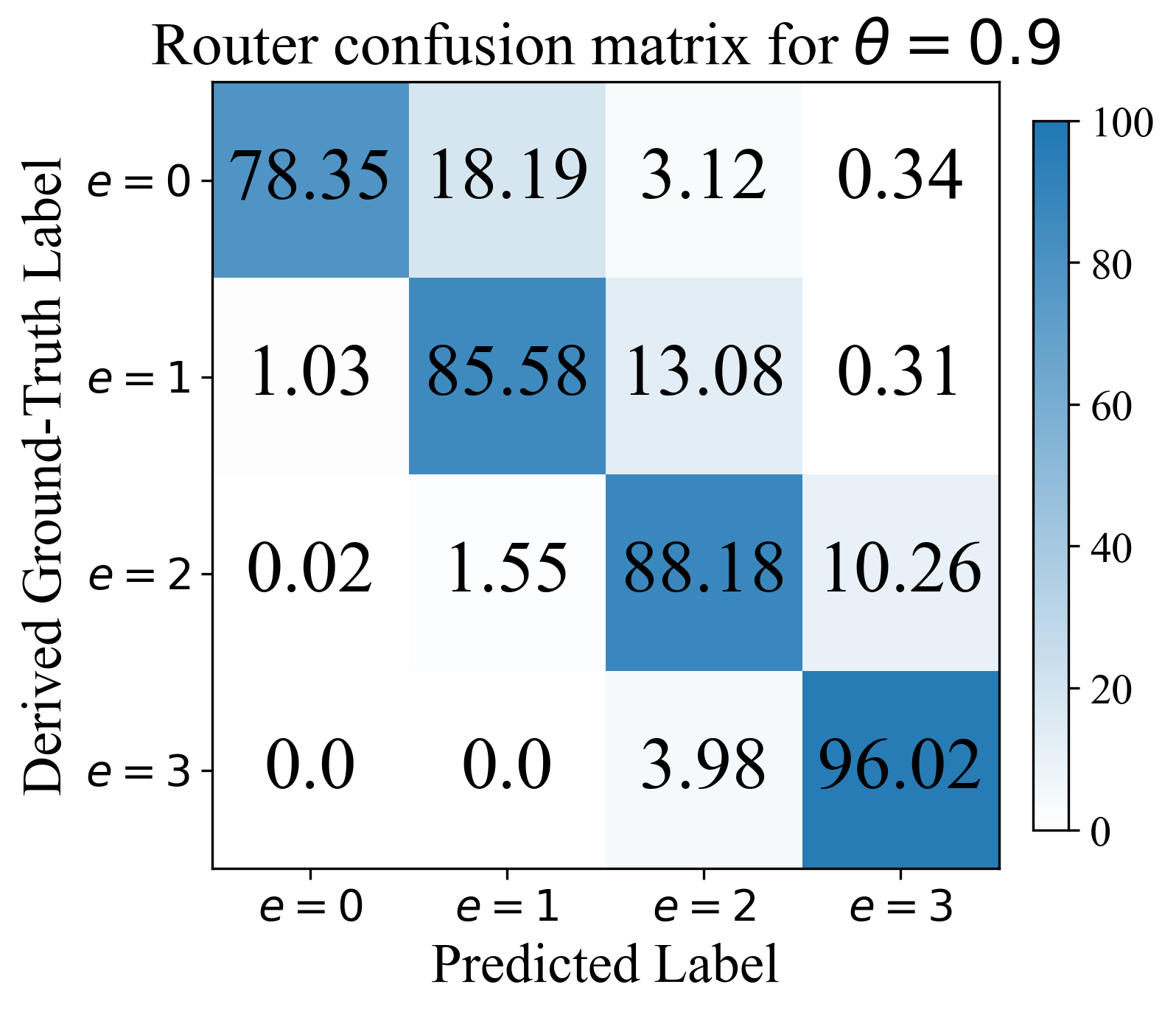}
         \caption{}
         \label{fig:0.9_routing}
     \end{subfigure}
     \caption{Confusion matrix for the router's classification task in $\mname$. The strong diagonal pattern in the matrices reflects high classification accuracy. The misclassified tokens often occurred in neighboring expert classes. 
     }
     \label{fig:router_confusion_matrix}     
\end{figure*}

\begin{figure*}[t!]
      \centering
    \begin{subfigure}[b]{0.48\linewidth}
         \centering
         \includegraphics[width=\linewidth]{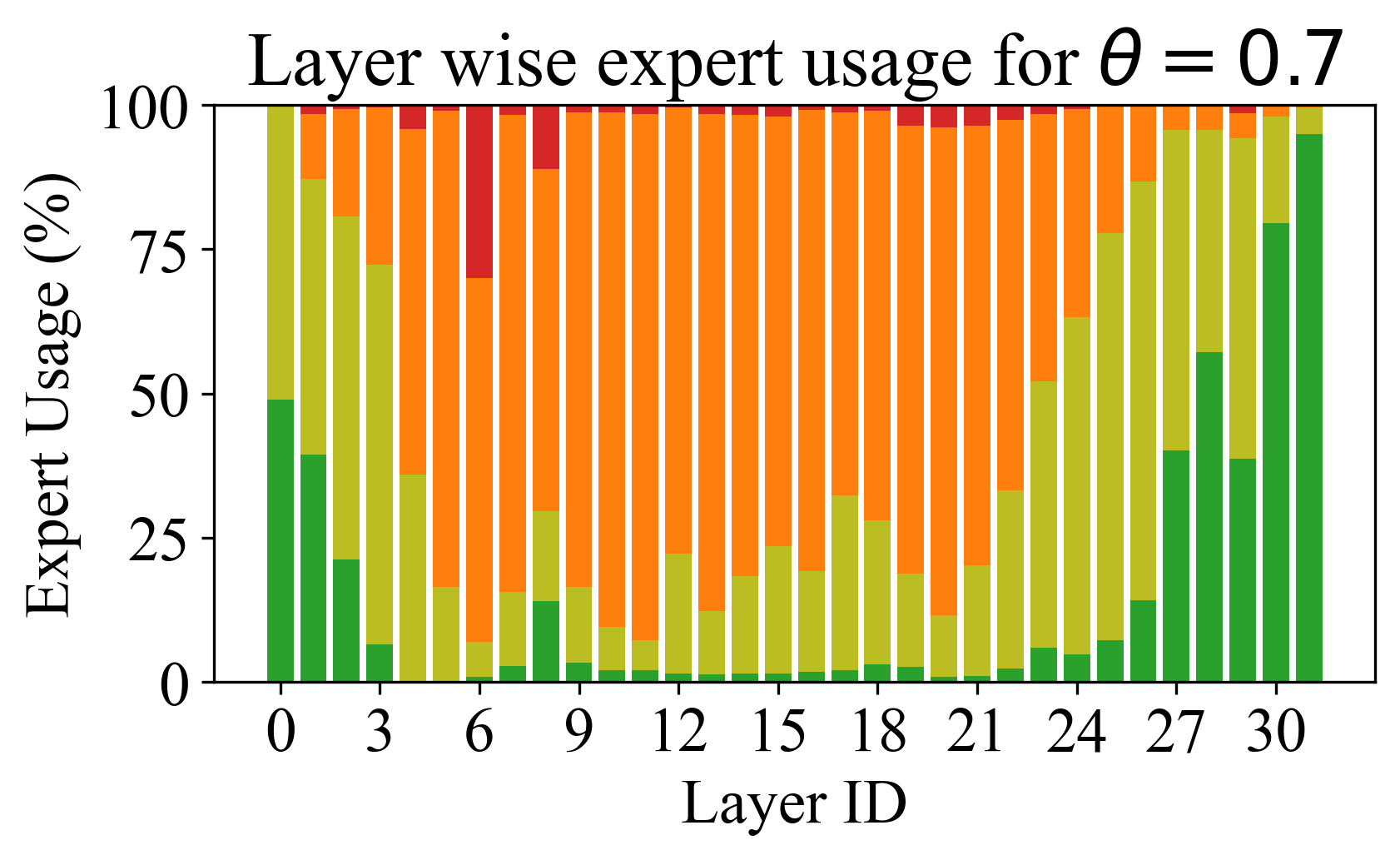}
         \caption{}
         \label{fig:0.7_gate}
     \end{subfigure}
     \begin{subfigure}[b]{0.48\linewidth}
         \centering
         \includegraphics[width=\linewidth]{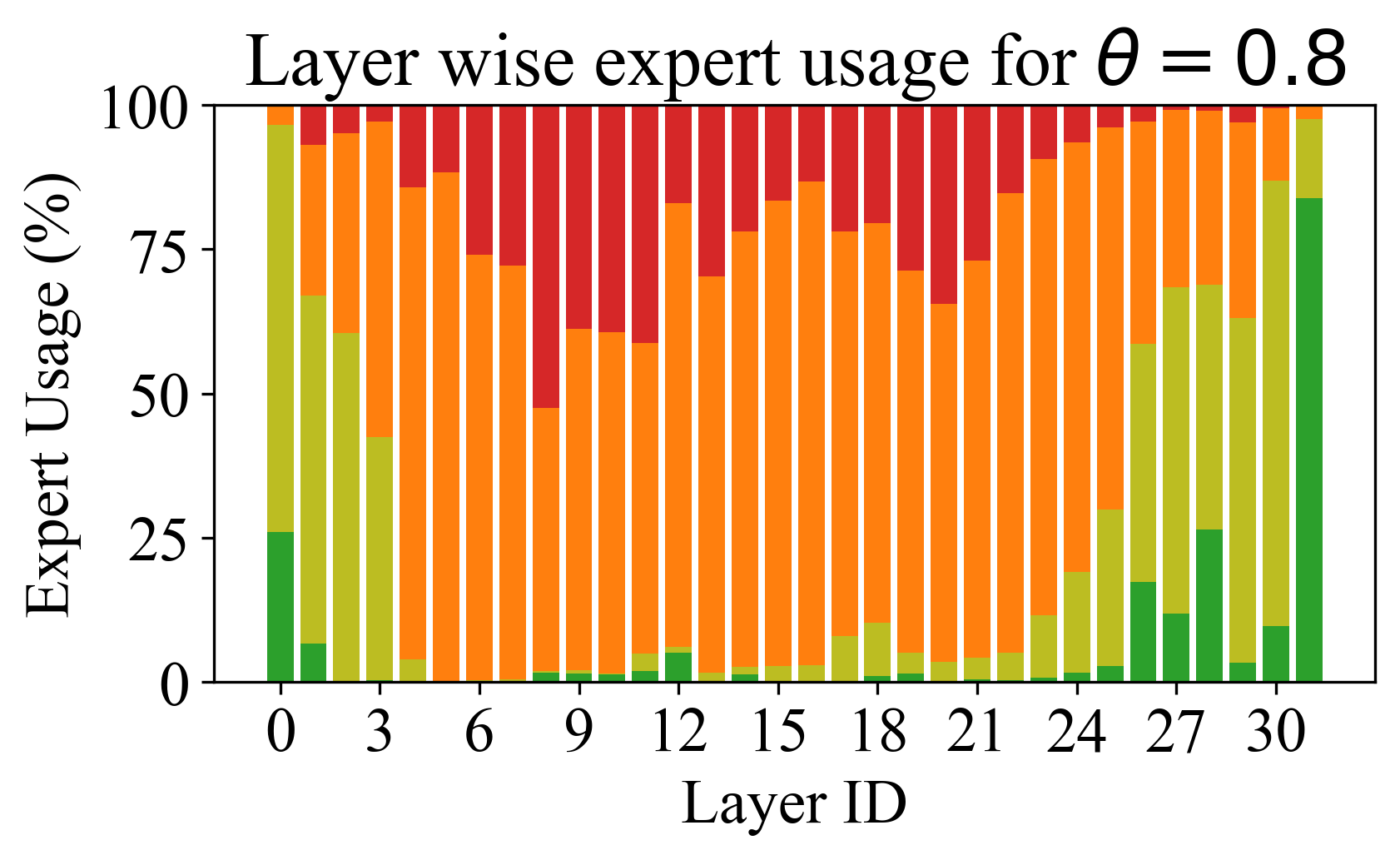}
         \caption{}
         \label{fig:0.8_gate}
     \end{subfigure}
    \begin{subfigure}[b]{0.48\linewidth}
         \centering
         \includegraphics[width=\linewidth]{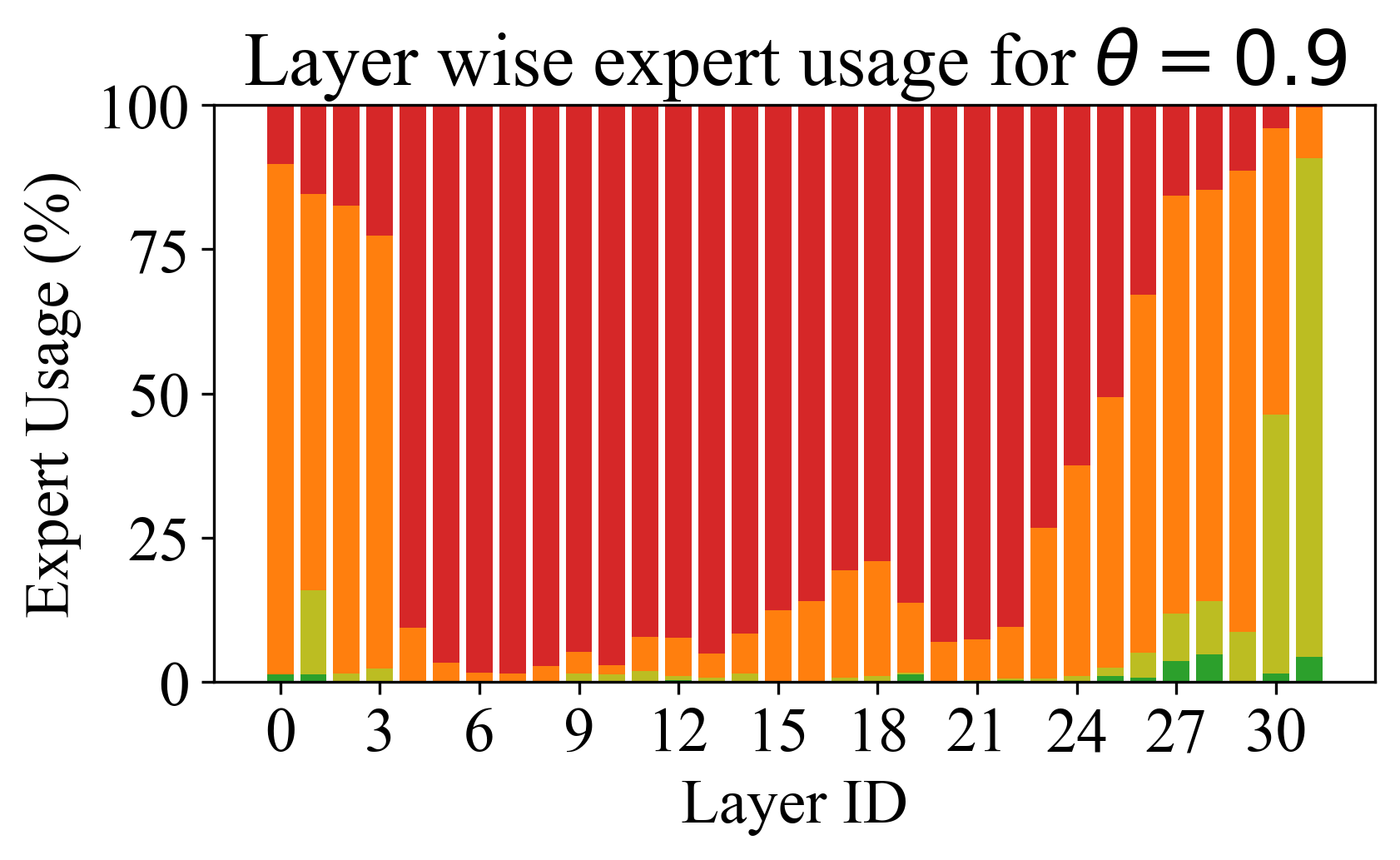}
         \caption{}
         \label{fig:0.9_gate}
    \end{subfigure}
    \begin{subfigure}[b]{0.48\linewidth}
         \centering
         \includegraphics[width=\linewidth]{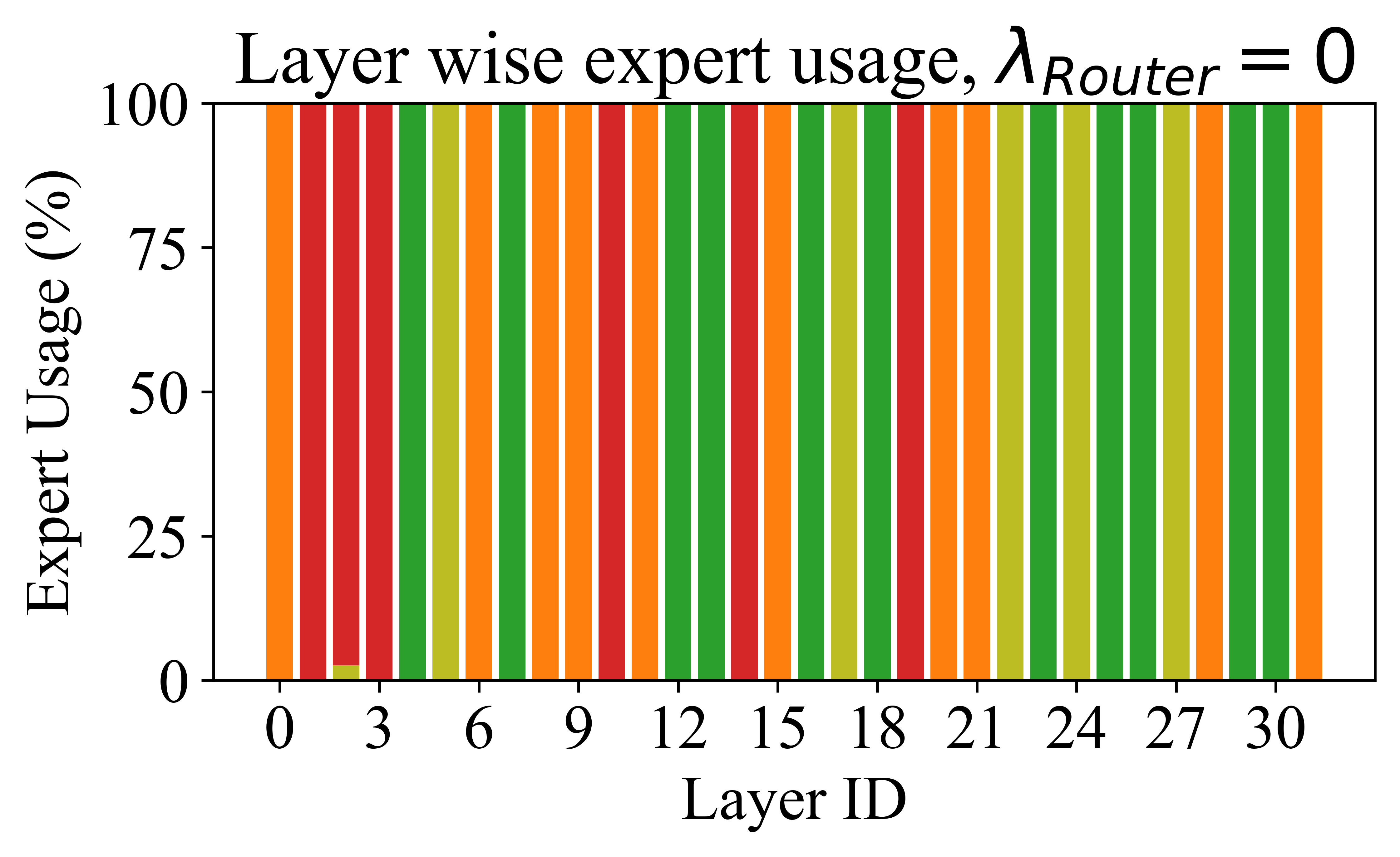}
         \caption{}
         \label{fig:nocam_gate}
    \end{subfigure}
    \begin{subfigure}[b]{0.9\linewidth}
         \centering
         \includegraphics[width=\linewidth]{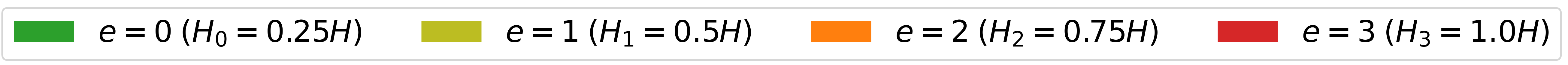}
    \end{subfigure}
         
     \caption{Layer-wise expert usage pattern of $\mname$ models with varying $\theta$, aggregated across $7$ downstream tasks. Each layer uses experts of varying sizes depending on the inputs. The sensitivity parameter $\theta$ regulates how readily tokens are routed to larger experts based on their difficulty: a lower $\theta$ favors smaller experts, while a higher $\theta$ prioritizes larger experts. \ref{fig:nocam_gate} shows the case when $\mname$ is trained without the router loss $\lambda_{Router}$. In this case, the model tends to assign a specific expert per layer instead of dynamically selecting experts based on token difficulty.
     }
     \label{fig:cam_expert_load}     
\end{figure*}

\section{Results}
\subsection{Evaluation}
\label{subsec:Evaluation}
We evaluate the $\mname$ models on $7$ downstream tasks using LM Evaluation Harness \cite{eval-harness} and report the 0-shot non-normalized accuracy metric in Table~\ref{tab:result}. The selected evaluation tasks include ARC (Easy and Challenge) \cite{arc-c-e}, HellaSwag \cite{Zellers2019HellaSwagCA}, PIQA \cite{Bisk2019PIQARA}, SciQ \cite{SciQ}, WinoGrande \cite{Winograd}, and LAMBADA \cite{lambda}.

$\mname$ is compared to two baselines, the base Mistral 7B model and the Flextron model ~\citep{Cai2024FlextronMF} using Avg4 and Avg7 in Table \ref{tab:result}. Compared to Mistral 7B, $\mname$ with $\theta=0.8$ improves efficiency by activating only $5.1B$ of $7B$ parameters on average, with an $7.3$ point accuracy drop after fine-tuning on only $10B$ tokens on the downstream tasks. The number of activated parameters adapts dynamically to token difficulty. For reference, Flextron fine-tunes on $93.57B$ tokens, activating $4.1B$ of $6.5B$ parameters, with a $5$ point accuracy drop from its base model, Llama2-7B \cite{LLM2_LLama}. We emphasize that with only $\frac{1}{9}\text{th}$ of the Flextron's fine-tuning cost, our results  for $\mname$ with $\theta=0.7$  are comparable to Flextron. Accuracy improves with increase in fine-tuning cost, but to keep the adaption lightweight, we opt for a smaller cost.


\subsection{Analysis of Token-Difficulty-Aware Router}
We assess the performance of the Token-Difficulty-Aware router by gathering its predictions from all layers across the 7 downstream tasks outlined in Section \ref{subsec:Evaluation}. These predictions are then compared to the ground truth labels derived in Section \ref{subsec:generate_gt_label}. Using both sets of labels, we compute the router's overall classification accuracy. 

We present the confusion matrices for the router's classification tasks across all $\mname$ models in Fig. \ref{fig:router_confusion_matrix}. Notably, the matrices exhibit a strong diagonal pattern, indicating high classification accuracy. Furthermore, when the router misclassifies tokens, the errors predominantly occurred in neighboring expert classes, underscoring the router's effectiveness in distinguishing token difficulty levels.

\subsection{Experts usage analysis}
We visualize the expert usage patterns across all layers in  Fig. \ref{fig:cam_expert_load}. For each model, the Y-axis represents the percentage of tokens routed to a specific expert, while the X-axis indicates the layer index. Notably, a token's perceived difficulty may vary across layers, hence it can be routed to different experts as the token progresses through the model. The visualization shows that all experts are utilized in varying proportions across layers, reflecting an aggregated behavior over the $7$ tasks. However, during inference, the model adapts to the data, with simpler queries predominantly engaging lower-compute experts to maximize efficiency.

The parameter $\theta$ affects expert usage in $\mname$ models by controlling how quickly tokens are routed to larger experts based on difficulty. At lower $\theta$ values (e.g., $\theta=0.7$), smaller experts ($e=2$) dominate across layers, optimizing for efficiency. In contrast, at higher $\theta$ values (e.g., $\theta=0.9$), larger experts ($e=3$) are utilized more frequently, prioritizing accuracy over efficiency. This shift demonstrates $\theta$'s role in balancing computational resource allocation and prediction accuracy.

\textbf{Ablating the Router Loss}: 
To examine the router loss's role in expert allocation, we train a $\mname$ model without it, relying solely on the LLM loss to train the router. Tokens are routed using the router’s predicted expert indices without explicit difficulty supervision. The resulting expert usage pattern, shown in Fig. \ref{fig:nocam_gate}, reveals that the model converges to using specific experts per layer instead of dynamically allocating experts based on token difficulty. In contrast, when router loss is applied, expert usage adapts dynamically to token difficulty across layers.




\section{Conclusion}
We present $\mname$, a post-training optimization framework that converts a standard pre-trained dense LLM into a token-difficulty-driven MoE model. $\mname$ incorporates a lightweight router to predict the token difficulty and routes them to an appropriate expert. To train this router, we propose a novel method to derive the token difficulty labels, which act as supervision signals. $\mname$ generates adaptive model variants with sensitivity control, allowing customization of the trade-off between efficiency and accuracy. 


\section*{Limitations}

While the proposed post-training optimization framework, $\mname$, and the token-difficulty-aware router provide a general-purpose approach for token-difficulty adaptive processing, this work does not explore their application to pre-trained heterogeneous mixture of expert (HMoE) models  \cite{Wang2024HMoEHM}. Such models, with their built-in experts of varying capabilities, could benefit from the integration of $\mname$, which provides a direct mechanism for routing tokens to appropriate experts based on token difficulty. Incorporating $\mname$ with HMoE models offers a promising direction for future exploration. Additionally, while $\mname$ makes the base model input-adaptive, offering inference efficiency, this work evaluates efficiency only in terms of the number of active parameters and does not include the efficiency measured in a real deployment scenario.



\bibliography{reference}


\end{document}